\documentclass{article}

\usepackage{arxiv}

\usepackage[utf8]{inputenc} 
\usepackage[T1]{fontenc}    
\usepackage{hyperref}       
\usepackage{url}            
\usepackage{booktabs}       
\usepackage{amsfonts}       
\usepackage{nicefrac}       
\usepackage{microtype}      
\usepackage{lipsum}

\usepackage{graphicx}
\usepackage{amssymb,amsmath,amsthm}

\usepackage{mdframed}
\usepackage{systeme,mathtools}
\usepackage{adjustbox}
\usepackage{hyperref}

\usepackage{comment}
\usepackage{todonotes}

\title{SafeML: Safety Monitoring of Machine Learning Classifiers through Statistical Difference Measure}

\author{
  Koorosh Aslansefat\\
  Department of Computer Science\\
  University of Hull\\
  Kingston upon Hull, HU6 7RX \\
  \texttt{k.aslansefat-2018@hull.ac.uk} \\
   \And
    Ioannis Sorokos \\
  Fraunhofer Institute for Experimental Software Engineering, IESE\\
  Fraunhofer-Gesellschaft\\
  Fraunhofer-Platz 1, 67663 Kaiserslautern, Germany \\
  \texttt{ioannis.sorokos@iese.fraunhofer.de} \\
     \And
    Declan Whiting \\
    Department of Computer Science\\
    University of Hull, APD Communications\\
    Kingston upon Hull, HU6 7RX, HU1 1RR \\
    \texttt{d.Whiting-2018@hull.ac.uk} \\
       \And
    Ramin Tavakoli Kolagari \\
    Faculty of Computer Science\\
    Nuremberg Tech\\
    Keßlerplatz 12, 90489 Nürnberg, Germany\\
    \texttt{ramin.tavakolikolagari@th-nuernberg.de} \\
         \And
    Yiannis Papadopoulos \\
    Department of Computer Science\\
    University of Hull\\
    Kingston upon Hull, HU6 7RX \\
    \texttt{y.i.papadopoulos@hull.ac.uk} \\
}

\begin{document}
\maketitle

\begin{abstract}
Ensuring safety and explainability of machine learning (ML) is a topic of increasing relevance as data-driven applications venture into safety-critical application domains, traditionally committed to high safety standards that are not satisfied with an exclusive testing approach of otherwise inaccessible black-box systems. Especially the interaction between safety and security is a central challenge, as security violations can lead to compromised safety. The contribution of this paper to addressing both safety and security within a single concept of protection applicable during the operation of ML systems is active monitoring of the behavior and the operational context of the data-driven system based on distance measures of the Empirical Cumulative Distribution Function (ECDF). We investigate abstract datasets (XOR, Spiral, Circle) and current security-specific datasets for intrusion detection (CICIDS2017) of simulated network traffic, using statistical distance measures including the Kolmogorov-Smirnov, Kuiper, Anderson-Darling, Wasserstein and mixed Wasserstein-Anderson-Darling measures. Our preliminary findings indicate that there is a meaningful correlation between ML decisions and the ECDF-based distances measures of the input features. Thus, they can provide a confidence level
that can be used for a) analyzing the applicability of the ML system in a given field (safety/security) and b) analyzing if the field data was maliciously manipulated\footnote{Our preliminary code and results are available at \href{https://github.com/ISorokos/SafeML}{https://github.com/ISorokos/SafeML}}.

\end{abstract}
\section{Introduction} \label{section-introduction}
Machine Learning (ML) is expanding rapidly in numerous applications. In parallel with this rapid growth, the expansion of ML towards dependability-critical applications raises societal concern regarding the reliability and safety assurance of ML. For instance, ML in medicine by \cite{begoli2019need,wiens2019no,qayyum2020secure}, in autonomous systems e.g. self-driving cars by \cite{burton2020mind,du2020ai}, in military \cite{sharkey2019autonomous}, and in economic applications by \cite{Davenport2019}. In addition, different organizations and governmental institutes are trying to establish new rules, regulations and standards for ML, such as in \cite{ISO_AI,UK_GOV,alexander2020safety}.

While ML is a powerful tool for enabling data-driven applications, its unfettered use can pose risks to financial stability, privacy, the environment and in some domains even life. Poor application of ML is typically characterized by poor design, misspecification of the objective functions, implementation errors, choosing the wrong learning process, or using poor or non-comprehensive datasets for training. Thus, safety for ML can be defined as a set of actions to prevent any harm to humanity by ML failures or misuse. However, there are many perspectives and directions to be defined for ML Safety. In fact, \cite{amodei2016concrete} have addressed different research problems of certifying ML systems operating in the field. They have categorized safety issues into five categories: a) safe exploration, b) robustness to distributional shift, c) avoiding negative side effects, d) avoiding “reward hacking” and “wire heading”, e) scalable oversight. This categorization is helpful for an adequate assessment of the applicability a concept for a given (safety) problem. In the work presented here, we will be focusing on addressing distributional shift, however using a non-standard interpretation. Distributional shift is usually interpreted as the gradual deviation of the initial state of learning of an ML component and its ongoing state as it performs online learning. As will be shown later, distributional shift will instead be used by our approach to evaluate the distance between the training and observed data of an ML component.

Statistical distance measures can be considered as a common method to measure distributional shift. Furthermore, in modern ML algorithms like Generative Adversarial Nets (GANs), statistical distance or divergence measures are applied as a loss function, such as the Jensen-Shannon divergence \cite{goodfellow2014generative}, the Wasserstein distance \cite{gulrajani2017improved}, and the Cramer distance \cite{bellemare2017cramer}. For dimension reduction, the t-SNE (t-distributed stochastic neighbour embedding) algorithm uses the Kullback-Leibler divergence as a loss function \cite{van2014accelerating}.

\subsection{Contributions}
This paper studies the applicability of safety-security monitoring based on statistical distance measures on the robustness of ML systems in the field.The basis of this work is a modified version of the statistical distance concept to allow the comparison of the data set during the ML training procedure and the observed data set during the use of the ML classifier in the field. The calculation of the distance is carried out in a novel controller-in-the-loop procedure to estimate the accuracy of the classifier in different scenarios. By exploiting the accuracy estimation, applications can actively identify situations where the ML component may be operating far beyond its trained cases, thereby risking low accuracy, and adjust accordingly. The main advantage of this approach is its flexibility in potentially handling a large range of ML techniques, as it is not dependent on the ML approach. Instead, the approach focuses on the quality of the training data and its deviation from the field data. In a comprehensive case study we have analyzed the possibilities and limitations of the proposed approach.

\subsection{Overview of the Paper}
The rest of the paper is organised as follows: In Section \ref{related-work}, previous work related to this publication is discussed.  In Section \ref{section-problem-definition}, the problem definition is provided. The proposed method is addressed in Section \ref{section-method}. Numerical results are demonstrated in Section \ref{section-case-studies} with a brief discussion. Explainable AI is introduced and discussed as a highly relevant topic in Section \ref{section-xai}. The capabilities and limitations of the proposed method are summarised in Section \ref{section-discussion} and the paper terminates with a conclusion in Section \ref{section-conclusion}. 

\subsection{Related Work} \label{related-work}
Our analysis of the research literature did not reveal any reference to existing publications dealing with the safety, security and accuracy of ML-based classification using statistical measures of difference. Nevertheless, there are publications that provide a basis for comparison with the current study.
A Resampling Uncertainty Estimation (RUE)-based algorithm has been proposed by \cite{schulam2019can} to ensure the point-wise reliability of the regression when the test or field data set is different from the training dataset. The algorithm has created prediction ensembles through the modified gradient and Hessian functions for ML-based regression problems. An uncertainty wrapper for black-box models based on statistical measures has been proposed by \cite{klas2019uncertainty}. Hobbhahn. M. et al. \cite{hobbhahn2020fast} have proposed a method to evaluate the uncertainty of Bayesian Deep Networks classifiers using Dirichlet distributions. The results were promising but to a limited class of classifiers (Bayesian Network-based classifiers). A new Key Performance Index (KPI), the Expected Odds Ratio (EOR) has been introduced in \cite{finlay2019empirical}. This KPI was designed to estimate empirical uncertainty in deep neural networks based on entropy theories. However, this KPI has not yet been applied to other types of machine learning algorithms. A comprehensive study on dataset shift has been provided by \cite{quionero2009dataset} and the dataset issues such as projection and projectability, simple and prior probability shift are discussed there. However, the mentioned study does not address the use of statistical distance and error bound to evaluate the dataset shift, in contrast to the work presented here.

\section{Problem Definition} \label{section-problem-definition}
\begin{figure}
\includegraphics[width=\textwidth]{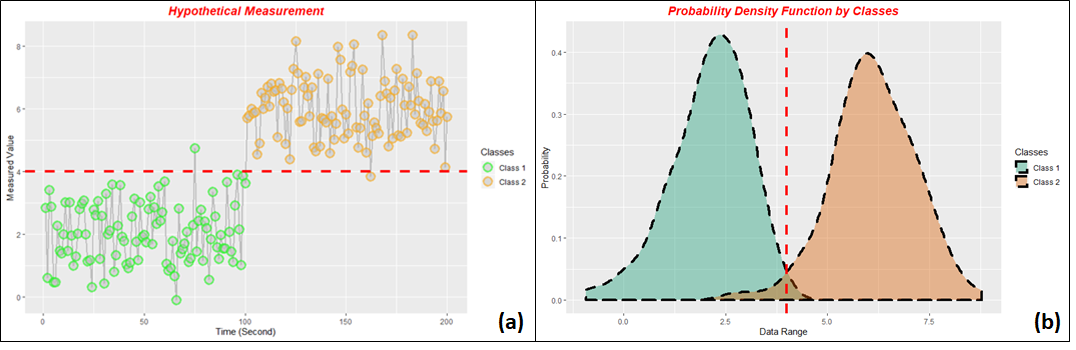}
\caption{(a) A hypothetical measurement (i.e. from 0 to 100 is Class 1 and from 101 to 200 is Class 2) (b) The estimated probability density function for both Class 1 and Class 2 with a classifier threshold equal to four} \label{fig1}
\end{figure}

Classification ML algorithms are typically employed to categorize input samples into predetermined categories. For instance, abnormality detection can be performed by detecting whether a given sample falls within known ranges i.e. categories. A simple example of a classifier for 1-dimensional input can be a line or threshold. Consider a hypothetical measurement t (e.g. time, temperature etc.) and a classifier D based on it, as shown in Figure \ref{fig1}-(a) and defined as (\ref{eq1}). Note that Figure \ref{fig1} shows the true classes of the input.
\begin{equation}
  D(t) = 
   \begin{cases} 
        Class1, & \mbox{if  } 0 < n \leq 100 \\  
        Class2, & \mbox{if  } 100 < n \leq 200 
   \end{cases}
   \label{eq1}
\end{equation}
The classifier $ D\left(t\right)\ $ can predict two classes which represent, in this example, the normal and abnormal state of a system. From measurement input 0 to 100, the sample is considered to fall under class 1 and from above 100 to 200 under class 2. The Probability Density Functions (PDFs) of the (true) classes can be estimated as shown in Figure \ref{fig1}-(b). In this figure, the threshold of the classifier has been represented with a red vertical dash-line and value of four. The area with an overlap in this figure can cause false detection to occur, as the classifier misclassifies the input belonging to the opposite class. These type of misclassifications are also known as false positive/type I errors (e.g. when misclassifying input as being class 1) and false-negative/type II errors (e.g. when misclassifying input as not being class 1).
Considering Figure \ref{fig1}-(b) of probability density functions, we notice that in the area where the two probability density functions merge, the misclassifications and thus the errors can occur. The probability of the error or misclassification can be calculated with (\ref{eq2}) \cite{Theodoridis2009}. Note that the error probability is also related to the threshold value (x considered as the threshold value), (for more details see \cite{aslansefat2020performance}).

\begin{equation}
    P\left(error\right)\ =\ \int_{-\infty}^{+\infty}P\left(error|x\right)P\left(x\right)dx
    \label{eq2}
\end{equation}

In listing (2), the $P\left(error|x\right)$ can be calculated as the minimum of both PDFs as (\ref{eq3}). The minimization is subject to variation of threshold value from $-\infty$ to $+\infty$.

\begin{equation}
    P\left(error|x\right)\ =\ min\left[P\left(Class\ 1|x\right),\ P\left(Class\ 2|x\right)\right]
    \label{eq3}
\end{equation}

By dividing the space into two regions as $R_1$ and $R_2$, the probability of error can be written in two parts.

\begin{equation} 
\begin{split}
    P\left(error\right)\ =\ P\left(x\in\ R_1,Class\ 1\right)+P\left(x\in R_2,Class\ 2\right) \\ =\int_{R_1} P\left(x|Class\ 1\right)P\left(Class\ 1\right)dx+\int_{R_2} P\left(x|Class\ 2\right)P\left(Class\ 2\right)dx
    \label{eq4}
    \end{split}
\end{equation}

To ease the minimization problem, consider the following inequality rule \cite{fukunaga2013introduction}.
    
\begin{equation} 
     min\left[a,b\right]\le a^\lambda b^{1-\lambda}\ where\ a,b\ \geq0\ and\ 0\le\alpha\le1
    \label{eq5}
\end{equation}

Equation (\ref{eq3}) can be rewritten as (\ref{eq6}). Note that in (\ref{eq5}) the $"\le"$ can considered as $"="$ when we consider the worst-case scenario or upper bound error.

\begin{equation}
\begin{split}
    P\left(error|x\right)\ =\ min\left[P\left(Class\ 1|x\right),\ P\left(Class\ 2|x\right)\right]= \\
    min\left[\frac{P\left(x|Class\ 1\right)P\left(Class\ 1\right)}{P\left(x\right)},\frac{P\left(x|Class\ 2\right)P\left(Class\ 2\right)}{P\left(x\right)}\right]
    \end{split}
    \label{eq6}
\end{equation} 

Using the inequality rule and equation (\ref{eq6}), the conditional probability of error can be derived as (\ref{eq7}).

\begin{equation}
\small
P\left(error|x\right)\ \le\left(\frac{P\left(x|Class\ 1\right)P\left(Class\ 1\right)}{P\left(x\right)}\right)^\lambda\left(\frac{P\left(x|Class\ 2\right)P\left(Class\ 2\right)}{P\left(x\right)}\right)^{1-\lambda}
\label{eq7}
\end{equation}

The equation (\ref{eq8}) can be obtained using equations (\ref{eq2}) and (\ref{eq7}).

\begin{equation} 
\begin{split}
    P\left(error\right)\ \le\left(P\left(Class\ 1\right)\right)^\lambda\left(P\left(Class\ 2\right)\right)^{1-\lambda}\ 
    \\
    \int_{-\infty}^{+\infty}{\left(P\left(x|Class\ 1\right)\right)^\lambda\left(P\left(x|Class\ 2\right)\right)^{1-\lambda}dx}
    \end{split}
    \label{eq8}
\end{equation}

In safety assurance, it is important to consider the worst-case scenario which can lead us to (\ref{eq9}), known as the Chernoff upper bound of error \cite{fukunaga2013introduction}.

\begin{equation}
\begin{split}
    P\left(error\right)\ = P\left(Class\ 1\right)^\lambda P\left(Class\ 2\right)^{1-\lambda}\
    \\
    \int_{-\infty}^{+\infty}{P\left(x|Class\ 1\right)^\lambda P\left(x|Class\ 2\right)^{1-\lambda}dx}
    \end{split}
    \label{eq9}
\end{equation} 

If the probability distributions of the features obey normal or exponential distribution families, the integral part of (\ref{eq9}) can be solved through (\ref{eq10}) \cite{fukunaga2013introduction}.

\begin{equation}
    \int_{-\infty}^{+\infty}{P\left(x|Class\ 1\right)^\lambda P\left(x|Class\ 2\right)^{1-\lambda}dx}=e^{-\theta\left(\lambda\right)}
    \label{eq10}
\end{equation} 

The $\theta\left(\lambda\right)$ can be calculated using (\ref{eq11}) where $\mu$ and $\Sigma$ are mean vector and variance matrix of each class respectively. 

\begin{equation}
\begin{split}
     \theta\left(\lambda\right)=\frac{\lambda\left(1-\lambda\right)}{2}\left[\mu_2-\mu_1\right]^T\left[\lambda\Sigma_1+\left(1-\lambda\right)\Sigma_2\right]^{-1}\left[\mu_2-\mu_1\right]
     \\
     +0.5\ log\frac{\left|\lambda\Sigma_1+\left(1-\lambda\right)\Sigma_2\right|}{\left|\Sigma_1\right|^\lambda\left|\Sigma_2\right|^{\left(1-\lambda\right)}}
    \label{eq11}
    \end{split}
\end{equation} 

Considering $\alpha=0.5$ the equation (\ref{eq11}) effectively becomes the Bhattacharyya distance. It can be proven that this value is the optimal value when $\Sigma_1=\Sigma_2$ \cite{fukunaga2013introduction,nielsen2018chord}. In this study, for simplicity, the Bhattacharyya distance will be used to demonstrate the approach. It should be noted that there may be cases where the calculated error bound is higher than the real value. However, this is acceptable as an overestimation of the classifier error would not introduce safety concerns (although it may impact performance). As the $P\left(error\right)$ and $P\left(correct\right)\ $ are complementary, the probability of having a correct classification can be calculated using (\ref{eq12}).

\begin{equation}
    P\left(correct\right)\ =1\ -\ \sqrt{P\left(Class\ 1\right)P\left(Class\ 2\right)}{\ e}^{-\theta\left(\lambda\right)}
    \label{eq12}
\end{equation} 

The Chernoff  upper bound of error is usually used as a measure of separability of two classes of data, but in the above context, equation (\ref{eq12}) measures the similarity between two classes. In other words, in an ideal situation, by comparing the $P\left(error\right)\ $ of a class, with itself, the response should be equal to one while $P\left(correct\right)$ should be zero. The intuitive explanation is to determine whether the distribution of the data during training is the same as the distribution observed in the field (or not). \\
Assuming $P\left(Class\ 1\right) = P\left(Class\ 2\right)$, the integral part of $P\left(error\right)$ can be converted to the cumulative distribution function as (\ref{eq13}).
\begin{equation}
\begin{split}
    P\left(error\right)\ =\ \left(\int_{-\infty}^{T}{P_{Class\ 1}\left(x\right)}dx+\int_{T}^{\infty}{P_{Class\ 2}\left(x\right)}dx\right)
    \\
    =\left(\int_{-\infty}^{T}{P_{Class\ 1}\left(x\right)}dx+\int_{T}^{+\infty}{P_{Class\ 2}\left(x\right)}dx\right)
    \\
    =\ \left(F_{Class\ 1}\left(T\right)+\left({1-F}_{Class\ 2}\left(T\right)\right)\right)
    \\
    =\ 1\ -\left(F_{Class\ 2}\left(T\right)-F_{Class\ 1}\left(T\right)\right)
    \end{split}
    \label{eq13}
\end{equation} 

Equation (\ref{eq13}) shows that there is relation between probability of error (and also accuracy) and statistical difference between two Cumulative Distribution Functions (CDF) of two classes. Using this fact and considering that the Empirical CDFs of each class is available, ECDF-based statistical measures such as  the Kolmogorov-Smirnov distance (equation \ref{eq14}) and similar distance measures can be used \cite{deza2014distances,raschke2011empirical}.

\begin{equation}
    P(error) \approx \sup_{x} { \left(F_{Class\ 2}\left(x\right)-F_{Class\ 1}\left(x\right)\right) }
    \label{eq14}
\end{equation}

It should be mentioned that such ECDF-based distances are not bounded between zero and one and, in some cases, need a coefficient to be adjusted as a measure for accuracy estimation. In section \ref{subsection-SHT}, the correlation between ECDF-based distance and accuracy will be discussed.

\section{SafeML Method} \label{section-method}
\begin{figure}
\includegraphics[width=\textwidth]{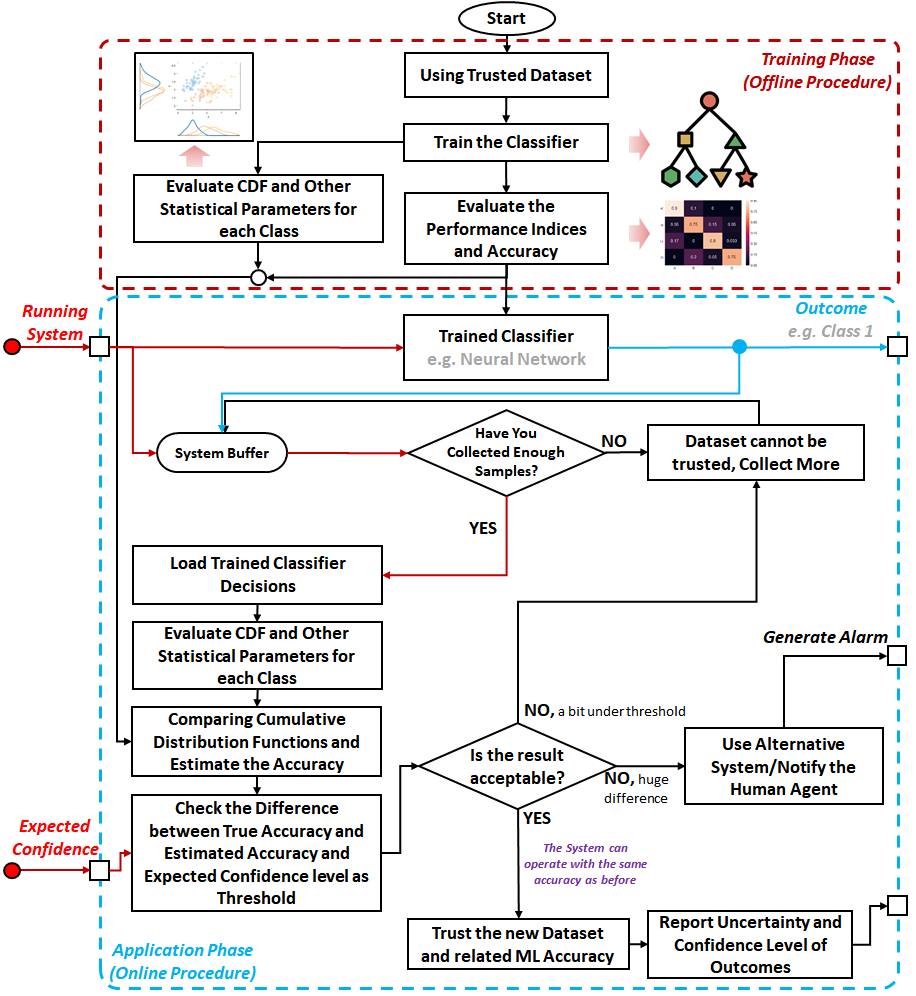}
\caption{Flowchart of the proposed approach} \label{fig2}
\end{figure}

To begin with, we should note that while this study focuses on ML classifiers, the proposed approach does not prohibit application on ML components for regression tasks either. Figure \ref{fig2} illustrates how we envision the approach to be applied practically. In this flowchart, there are two main sections; the training phase and the application phase. 
A) The 'training' phase is an offline procedure in which a trusted dataset is used to train the ML algorithm.  Once training is complete, the classifier's performance is measured with user-defined KPIs. Meanwhile, the PDF and statistical parameters of each class are also computed and stored for future comparison in the second phase. 
B) The second or 'application' phase is an online procedure in which real-time and unlabelled data is provided to the system. For example, consider an autonomous vehicle's machine vision system. Such a system has been trained to detect obstacles (among other tasks), so that the vehicle can avoid collisions with them. A critical issue to note in the application phase is that the incoming data is unlabeled. So, it cannot be assured that the classifier will perform as accurately as it had in during the training phase. As input samples are collected, the PDF and statistical parameters of each class can be estimated. The system requires enough samples to reliably determine the statistical difference, so a buffer of samples may have to be accumulated before proceeding. Using the modified Chernoff error bound in \ref{eq12},  the statistical difference of each class in the training phase and application phase is compared. If the statistical difference is very low, the classifier results and accuracy can be trusted. In the example mentioned above, the autonomous vehicle would continue its operation in this case.  Instead, if the  statistical difference is greater, the classifier results and accuracy are no longer considered valid (as the difference between the training and observed data is too large). In this case, the system should use an alternative approach or notify a human operator. In the above example, the system could ask the driver to takeover control of the vehicle.

\section{Case Studies} \label{section-case-studies}
\begin{figure}
\includegraphics[width=\textwidth]{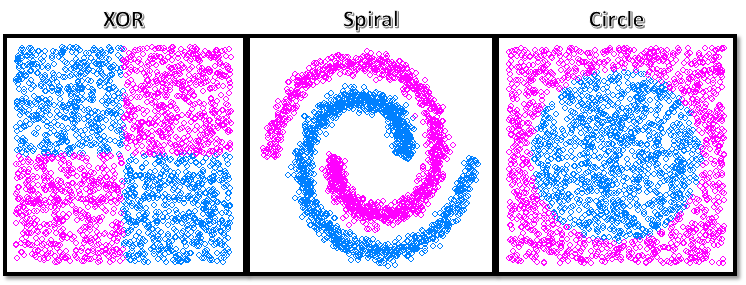}
\caption{Scatter plot of XOR, Spiral and Circle Benchmarks} \label{fig_Bench}
\end{figure}

In this section, the proposed method described in Section \ref{section-method} is applied on typical synthetic benchmarks for ML classification.
The proposed method has been implemented in three different programming languages including R, Python and MATLAB. Regarding R programming, three well-known benchmarks have been selected: a) the XOR dataset, b) the Spiral dataset and c) the Circle dataset. Each dataset has two features (i.e. input variables) and two classes. Figure \ref{fig_Bench} illustrates the scatter plots of the selected benchmarks. More examples and benchmarks are available at \href{https://github.com/ISorokos/SafeML}{SafeML Github Repository}.

\subsection{Methodology for Evaluation against Benchmark Datasets}

To start the ML-based classification, 80 percent of each dataset was used for training and testing and 20 percent of the dataset has been used for validation, with 10-fold cross-validation. Both linear and nonlinear classifiers have been selected for classification. The Linear discriminant analysis (LDA) and the Classification And Regression Tree (CART) are used as linear methods. Moreover, The Random Forest (RF), K-Nearest Neighbours (KNN) and Support Vector Machine (SVM) are applied as nonlinear methods.  As KPIs, the accuracy and Kappa measure are used to measure the performance of each classifier. Finally, as Empirical Cumulative Distribution Function (ECDF)-based statistical distance measures, the Kolmogorov-Smirnov Distance (KSD), Kuiper Distance, Anderson-Darling Distance (ADD), Wasserstein Distance (WD), and a combination of ADD and Wasserstein-Anderson-Darling Distance (WAD) have been selected for evaluation. 

\textbf{XOR Dataset:}
The XOR dataset has two features and two classes in which features have the same mean and variance characteristics. Table \ref{Tb1l} compares the estimated accuracy based on the ECDF measures with the Minimum True Accuracy (MTA) and the Average True Accuracy (ATA) over 10 folds. For instance, the second column of this table provides the estimated accuracy based on the KSD measure. As a matter of safety, MTA is more important because it represents the worst-case scenarios, where the lowest accuracy may be experienced and impact safety. We observe that the KSD measure reports low accuracy for the LDA classifier (~.77). Instead, the ADD and WAD measures significantly overestimate the accuracy of the LDA.

\begin{table}[htbp]
  \centering
  \caption{Comparison of estimated accuracies vs minimum true accuracy for XOR dataset}
  \scalebox{0.85}{
    \begin{tabular}{lcccccccccccccccc}
    \hline \hline
    Method   &    & KSD      &    & Kuiper  &    & ADD      &    & WD       &    & WAD      &    & BD       &    & MTA      &    & ATA \\
    \hline
    LDA      &          & 0.772217 &          & 0.770600   &          & 0.902818 &          & 0.755064 &          & 0.985666 &          & 0.154506 &          & 0.50833 &          & 0.59121 \\
    CART     &          & 0.928179 &          & 0.921982 &          & 0.987722 &          & 0.92545 &          & 0.995211 &          & 0.497243 &          & 0.98744 &          & 0.99415 \\
    KNN      &          & 0.93057 &          & 0.913063 &          & 0.993151 &          & 0.958768 &          & 0.997076 &          & 0.497102 &          & 0.97489 &          & 0.98666 \\
    SVM      &          & 0.931045 &          & 0.917586 &          & 0.993489 &          & 0.95819 &          & 0.997064 &          & 0.496731 &          & 0.97916 &          & 0.98791 \\
    RF       &          & 0.92962 &          & 0.910749 &          & 0.992742 &          & 0.957821 &          & 0.997018 &          & 0.496856 &          & 0.99583 &          & 0.99833 \\
    \hline \hline
    \end{tabular}}
  \label{Tb1l}%
\end{table}%

Based on Table \ref{Tb1l}, Table \ref{TB_XOR} represents the (absolute) difference between accuracy estimations of each measure and the MTA of each classifier. The ADD, WD and WAD measures have the best accuracy estimations overall. In particular, when a LDA classifier is used, the WD measure provides an estimated accuracy with comparatively less error.

\begin{table}[htbp]
  \centering
  \caption{Difference between Distance Measures and MTA for XOR dataset}
    \begin{tabular}{lrcrcrcrcrcrr}
      \hline \hline
    Method   &          & KSD      &          & Kuiper  &          & ADD      &          & WD       &          & WAD      &          & \multicolumn{1}{c}{BD} \\
      \hline 
    LDA      &          & 0.263883 &          & 0.262267 &          & 0.394484 &          & 0.246731 &          & 0.477333 &          & 0.353828 \\
    CART     &          & 0.059269 &          & 0.065466 &          & 0.000274 &          & 0.06199  &          & 0.007763 &          & 0.490205 \\
    KNN      &          & 0.044320  &          & 0.061833 &          & 0.018256 &          & 0.016127 &          & 0.02218  &          & 0.477793 \\
    SVM      &          & 0.048122 &          & 0.061580  &          & 0.014322 &          & 0.020976 &          & 0.017897 &          & 0.482310 \\
    RF       &          & 0.066207 &          & 0.085084 &          & 0.003092 &          & 0.038012 &          & 0.001184 &          & 0.499102 \\
      \hline \hline
    \end{tabular}%
  \label{TB_XOR}%
\end{table}%

\textbf{Spiral Dataset:} Similar to the XOR dataset, the proposed method can be applied for the spiral dataset. Table \ref{TB_Spiral} presents difference between ECDF-based distance measures and minimum true accuracy for this dataset. For brevity, for this dataset and the next one, only the difference table is provided. Based on this table, the KSD and Kuiper distance have better estimation for accuracy of the classifiers for the spiral dataset. 

\begin{table}[htbp]
  \centering
  \caption{Difference between Distance Measures and MTA for Spiral dataset}
    \begin{tabular}{lrcrcrcrcrcrc}
    \hline \hline
    Method   &          & KSD      &          & Kuiper &          & ADD      &          & WD       &          & WAD      &          & BD \\
    \hline 
    LDA      &          & 0.099447 &          & 0.088252 &          & 0.269975 &          & 0.248396 &          & 0.528852 &          & 0.043445 \\
    CART     &          & 0.056131 &          & 0.031092 &          & 0.149191 &          & 0.09477  &          & 0.158529 &          & 0.355675 \\
    KNN      &          & 0.047526 &          & 0.075598 &          & 0.001468 &          & 0.014756 &          & 0.002734 &          & 0.496559 \\
    SVM      &          & 0.047526 &          & 0.075598 &          & 0.001468 &          & 0.014756 &          & 0.002734 &          & 0.496608 \\
    RF       &          & 0.024471 &          & 0.050261 &          & 0.018778 &          & 0.003885 &          & 0.019643 &          & 0.479893 \\
    \hline \hline
    \end{tabular}%
  \label{TB_Spiral}%
\end{table}%

\textbf{Circle dataset:} The circle dataset has similar statistical characteristics with the spiral dataset. Table \ref{TB_Circle} provides the difference between ECDF-based distance measures and MTA for this dataset. As can be seen, the worst accuracy estimation is related to the accuracy estimation of the LDA classifier. For the LDA, the Kuiper distance estimates with less error, with the KSD and WD being in second and third place respectively.
\begin{table}[htbp]
  \centering
  \caption{Difference between Distance Measures and MTA for Circle dataset}
    \begin{tabular}{lrcrcrccccccc}
    \hline \hline
    Method   &          & KSD      &          & Kuiper  &          & ADD      &          & WD       &          & WAD      &          & BD \\
    \hline 
    LDA      &          & 0.329391 &          & 0.250345 &          & 0.412382 &          & 0.347450  &          & 0.49826 &          & 0.236670 \\
    CART     &          & 0.114312 &          & 0.019111 &          & 0.168596 &          & 0.099549 &          & 0.24322 &          & 0.455675 \\
    KNN      &          & 0.004833 &          & 0.037554 &          & 0.027649 &          & 0.010871 &          & 0.02775  &          & 0.498459 \\
    SVM      &          & 0.016133 &          & 0.043604 &          & 0.019147 &          & 0.001695 &          & 0.01935  &          & 0.498808 \\
    RF       &          & 0.004663 &          & 0.034529 &          & 0.027776 &          & 0.012814 &          & 0.02782  &          & 0.468893 \\
    \hline \hline
    \end{tabular}%
  \label{TB_Circle}%
\end{table}%

\subsection{Security dataset} \label{case-study-security}

\begin{figure}
\centering
\includegraphics[width=0.8\textwidth]{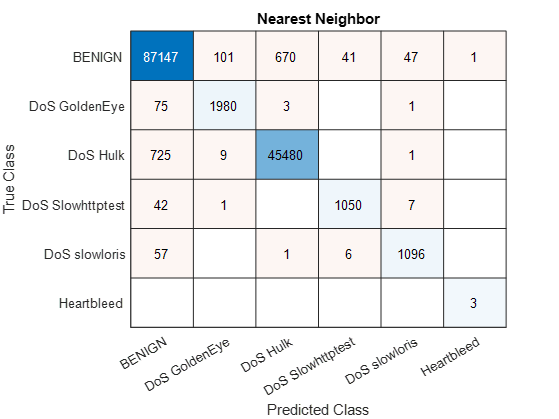}
\caption{Confusion matrix for Wednesday Security Intrusion Detection in CICIDS2017 Dataset} \label{fig_W01}
\end{figure}

This case-study applies the proposed method towards the CICIDS2017 dataset, which was originally produced by \cite{securitydataset} at the Canadian Institute for Cyber Security (CICS) as an aide to the development and research of anomaly-based intrusion detection techniques for use in Intrusion Detection Systems (IDSs) and Intrusion Prevention Systems (IPSs) \cite{datasetanalysis}.
The labelled dataset includes both benign (Monday) and malicious (Tuesday, Wednesday, Thursday, Friday) activity. The benign network traffic is simulated by abstraction of typical user activity using a number of common protocols such as HTTP, HTTPS, FTP and SHH. Benign and malicious network activity is included as packet payloads in packet capture format (PCAPS).
\\
\textbf{Wednesday Attack:} This attack occurred on Wednesday, July 5, 2017, and different types of attacks on the availability of the victim's system have been recorded, such as DoS / DDoS, DoS slowloris (9:47 – 10:10 a.m.), DoS Slowhttptest (10:14 – 10:35 a.m.), DoS Hulk (10:43 – 11 a.m.), and DoS GoldenEye (11:10 – 11:23 a.m.). Regarding the cross-validation, a hold-out approach has been used, in which 70 percent of data has been randomly extracted for testing and training and the rest has been used for accuracy estimation. Additionally, traditional classifiers including 'Naive Bayes','Discriminant Analysis','Classification Tree', and 'Nearest Neighbor' have been used. Figure \ref{fig_W01} shows the confusion matrix when Naive Bayes classifier is used. 

\begin{figure}
\includegraphics[width=\textwidth]{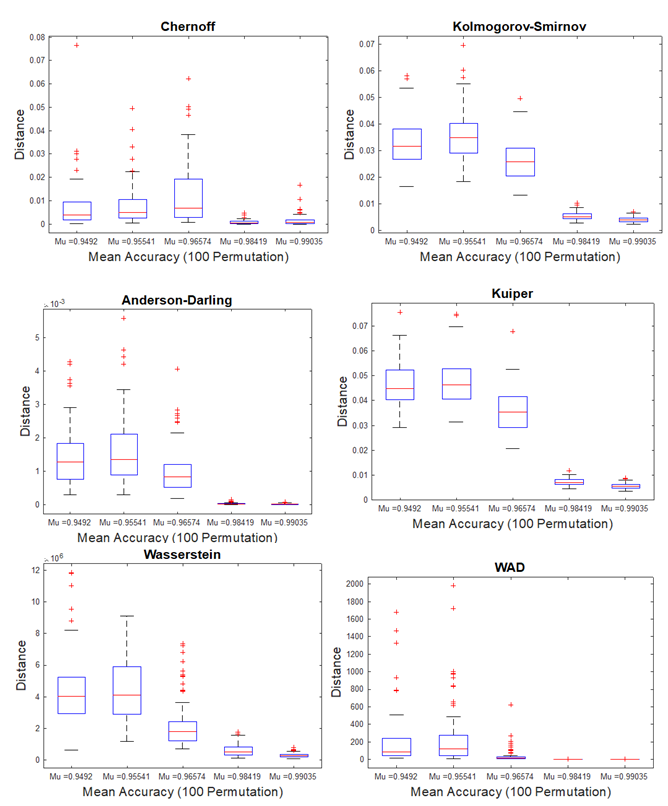}
\caption{Box plot of statistical distance measures vs. accuracy over 100 iterations} \label{fig_W03}
\end{figure}

Figure \ref{fig_W03} has been generated over 100 iterations. For each iteration, 70 percent of the data has been randomly extracted for testing and training and the rest has been used for accuracy estimation. Figure \ref{fig_W03} shows the box plot of the statistical distance measurements vs. the evaluated accuracy over 100 iterations. By observing the average values (red lines) of each box plot, the relationship between each measure and the average change in accuracy can be understood. In addition, this plot shows which method has less variation. For instance, the Kuiper distance and WD have the best performance while Chernoff has the least performance.

\begin{figure}
\centering
\includegraphics[width=0.8\textwidth]{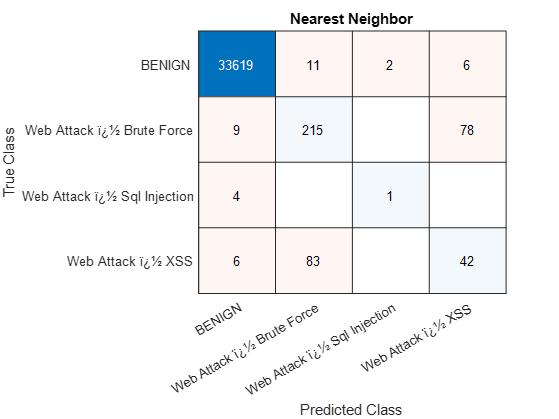}
\caption{Confusion matrix for Thursday Security Intrusion Detection in CICIDS2017 dataset} \label{fig_Th01}
\end{figure}

\textbf{Thursday Attack:} This attack occurred on Thursday, July 6, 2017, and various attacks, such as the Web Attack – Brute Force (9:20 – 10 a.m.), Web Attack – XSS (10:15 – 10:35 a.m.), and Web Attack – Sql Injection (10:40 – 10:42 a.m.) have been recorded. Figure \ref{fig_Th01} shows the confusion matrix for Thursday morning's security intrusion in the CICIDS2017 dataset when the Naive Bayes classifier is applied. Similar to Wednesday, 70 percent hold-out cross validation is used for this dataset. As can be seen, this dataset has four classes and the classifier has problem to detect the last class or last type of intrusion. 

\begin{figure}
\includegraphics[width=\textwidth]{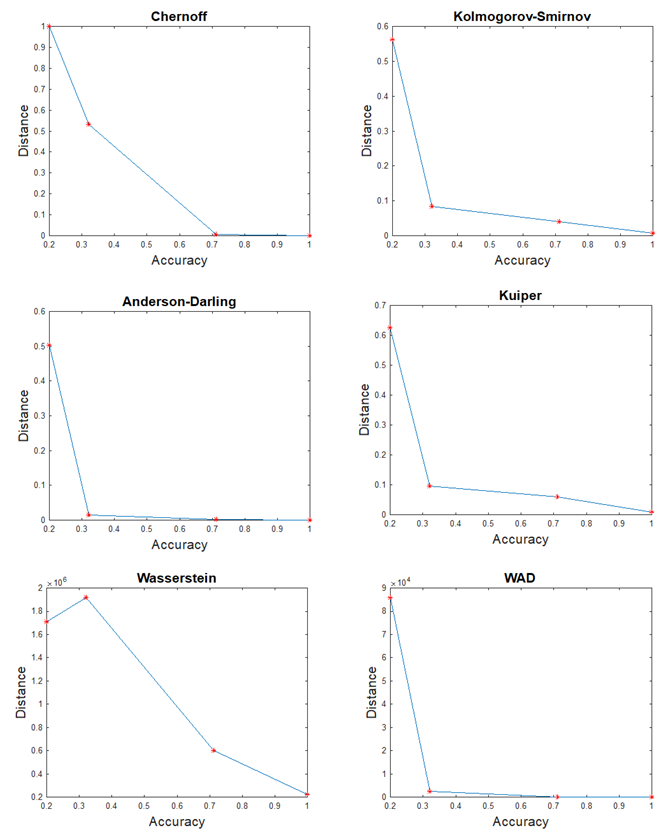}
\caption{Sample statistical distance measures vs. accuracy for Thursday Security Intrusion Detection in CICIDS2017 dataset} \label{fig_Th02}
\end{figure}

Figure \ref{fig_Th02} shows a sample result of six statistical measures (Chern-off and five ECDF-based measures) vs. accuracy of the classifier. In this sample, the Kolmogorov-Smirnov and Kuiper measures have better performance.

\begin{figure}
\includegraphics[width=\textwidth]{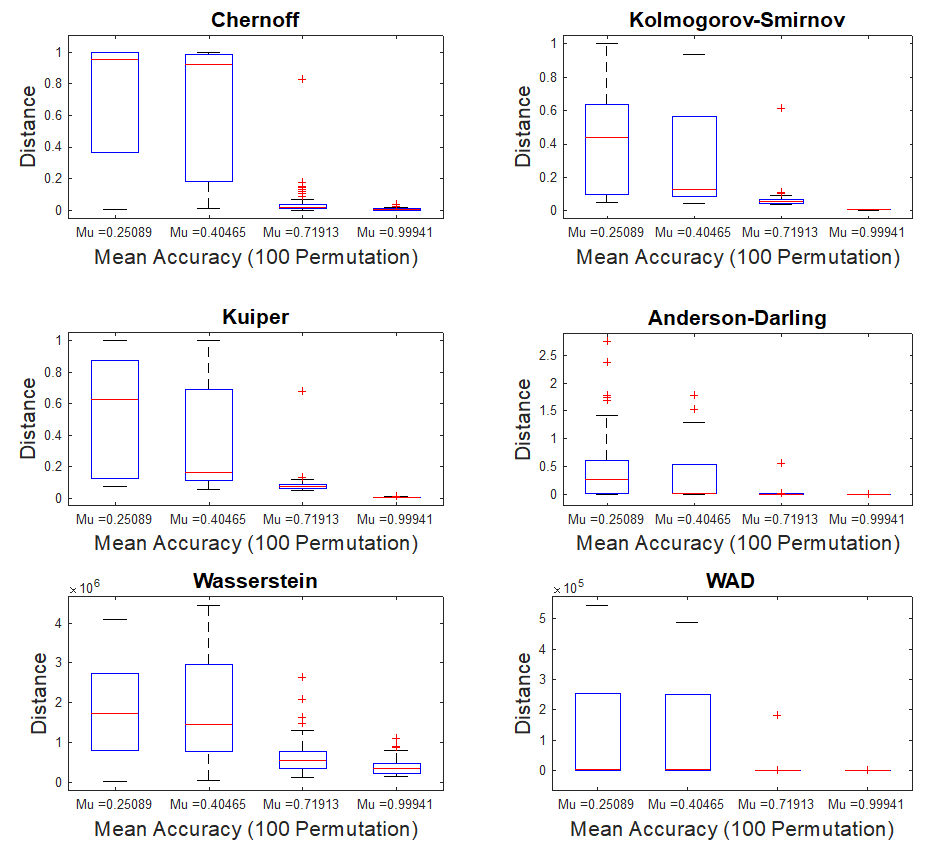}
\caption{Box plot of statistical distance measures vs. accuracy over 100 iterations for Thursday Security Intrusion Detection in CICIDS2017 dataset} \label{fig_Th03}
\end{figure}

Similar to the previous example, Figure \ref{fig_Th02} has been generated over 100 times and the box plot of Figure \ref{fig_Th03} can be seen. In this figure, the Kolmogorov-Smirnov, Kuiper and Wassertein distance measures have a better performance, however, their decision variance is a bit high.

The rest of results for Security Intrusion Detection in CICIDS2017 dataset are available in the \href{https://github.com/ISorokos/SafeML}{SafeML Github Repository}. 


\subsection{Correlation Analysis} \label{subsection-SHT}
\begin{figure}
\centering
\includegraphics[width=0.7\textwidth]{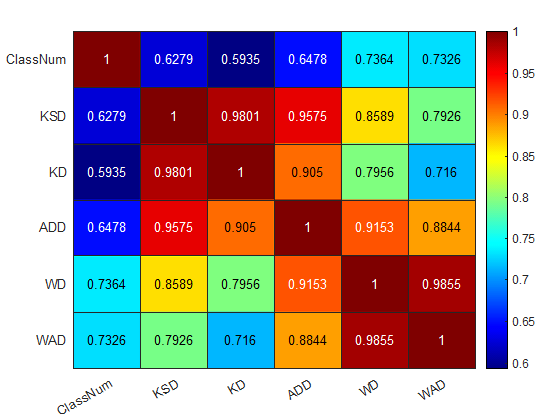}
\caption{Correlation between class label numbers and statistical ECDF-based Distance Measures} \label{fig_corr}
\end{figure}

Figure \ref{fig_corr} shows Pearson's correlation between the classes of Wednesday's data and the statistical ECDF-based distances. As can be seen, the WD and WAD distances have more correlation with the classes. This figure also shows the correlation between the measures themselves. The KSD and KD appear to be correlated. The WD and WAS also seem to be correlated. These correlations can be explained due to the similarity in their formulation. P-values for the above correlations were evaluated to be zero, thereby validating the correlation hypotheses above.

\section{Towards Explainable AI} \label{section-xai}
\begin{figure}
\includegraphics[width=0.95\textwidth]{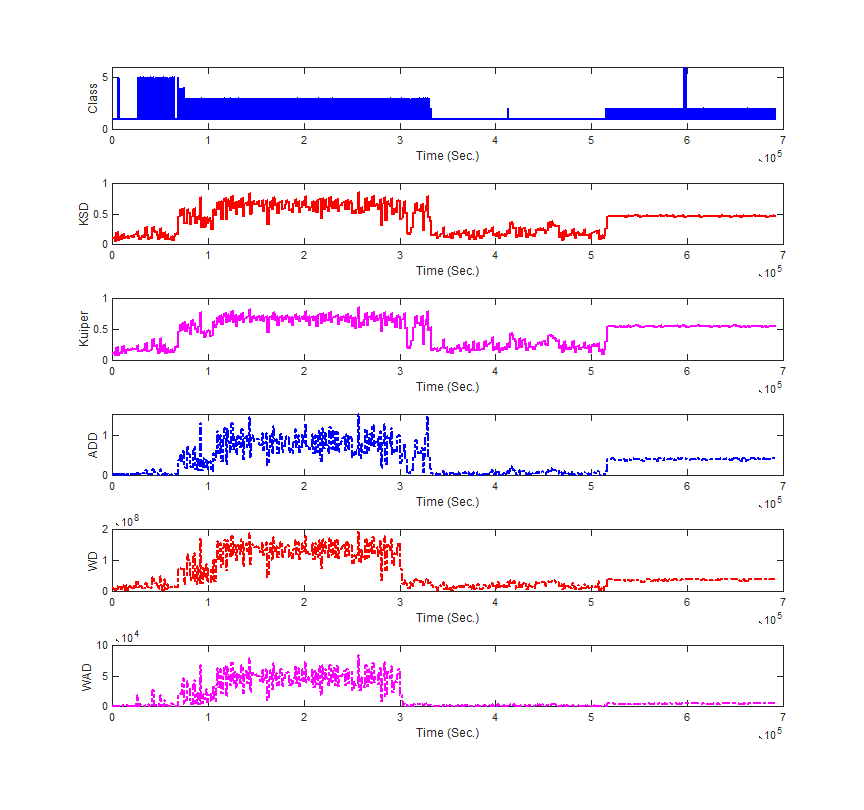}
\caption{Plot of class label and statistical ECDF-based Distances vs. time (Security dataset: Wednesday)} \label{fig_XAI}
\end{figure}

In this section, we discuss a relevant topic to our proposed approach, to explain how the proposed approach could be applied for this purpose as well. Explainable AI (XAI) can be defined as a tool or framework that increases interpretability of ML algorithms and their outputs \cite{cutillo2020machine}. Our proposed approach can also be used to improve the interpretability of ML classifiers using the statistical ECDF-based distance measures seen previously. We shall discuss a small example here and intend to delve further on this topic in our future works. For the example, the Wednesday data from the security dataset mentioned previously is chosen and its class labels vs. the sample time has been plotted in Figure \ref{fig_XAI}. This dataset has six different classes with variable number of occurrence. In this figure a sliding window with the size of $d=1500$ is used. In the beginning, $1500$ samples of class one are considered as reference and then compared with the rest of the samples for each window using the statistical ECDF-based distance measures. It should be mentioned that the smoothness of the output is related to the sliding window's size. As can be see in the figure, the change in the average distance vs. the class shows the existing high correlation. In addition, it seems that class number five is slightly robust to statistical change and class number six has a low number of samples, that cannot produce meaningful statistical difference. The problem of detecting class six can be solved by decreasing the size of the sliding window.This figure can be generated for different classifiers and show how their decisions are correlated to the ECDF-based distance measures. As an future work, we aim to investigate ECDF-based distance inside different algorithms to better understand their actions. This section is just a hint for future works. 

\section{Discussion} \label{section-discussion}
Overall, our preliminary investigation indicates that statistical distance measures offer the potential for providing a suitable indicator for ML performance, specifically for accuracy estimation. In particular, we further denote the following capabilities and limitations for the proposed approach.
\subsection{Capabilities of SafeML}
\begin{itemize}
 \item By modifying the existing statistical distance and error bound measures, the proposed method enables estimation of the accuracy bound of the trained ML algorithm in the field with no label on the incoming data.
 \item A novel human-in-loop monitoring procedure is proposed to certify the ML algorithm during operation. The procedure has three levels of operation: I) nominal operation allowed with assured ML-accuracy based on the distance estimation, II) buffering data samples to generate estimation, and III) low estimated accuracy estimated, leading to external intervention by automated/human controller being needed. 
 \item The proposed approach is easy to implement, and can support a variety of distributions (i.e. exponential and normal distribution families).
 \item The outcome of the proposed approach can be used as an input for runtime safety analysis in adaptive systems \cite{kabir2019runtime,papadopoulos2019model}
\end{itemize}
\subsection{Limitations of the proposed method}
\begin{itemize}
\item The proposed algorithm is currently only tackling the safety evaluation problem of the machine-learning-based classification. However, we believe it can be easily expanded for clustering, dimension reduction or any problem that can be evaluated through statistical difference.
\item Some of the machine learning algorithms can be robust to a certain distributional shift or variation in the dataset distribution. This may limit the effectiveness of the discussed distance measures. That being said, the proposed measures can then be used as additional confirmation of the robustness, contributing to certification arguments.
\end{itemize}

\section{Conclusion} \label{section-conclusion}
The expansion of ML applications to safety-critical domains is a major research question. We investigate the problem of context applicability of an ML classifier, specifically the distributional shift between its training and observed data. We have identified and evaluated sets of statistical distance measures that can provide estimated upper error bounds in classification tasks based on the training and observed data distance. Further, we have proposed how this approach can be used as part of safety and security-critical systems to provide active monitoring and thus improve their robustness. The overall most effective distance measure was identified to be the Kolmogorov-Smirnov. The proposed human-in-the-loop procedure uses this statistical distance measure to monitor the estimated accuracy of the ML component and notify its AI or human controller when the deviation exceeds specific boundaries.  The study is still in its early stages, but we believe the results to offer a promising starting point. The strengths and weaknesses of the proposed approach are discussed in the previous section.
\nocite{*}
\section*{Acknowledgements} \label{section-acknowledgements}
This work was supported by the DEIS H2020 Project under Grant 732242. We would like to thank EDF Energy R$\&$D UK Centre, AURA Innovation Centre and the University of Hull for their support.

\bibliographystyle{unsrt} 

\begin{thebibliography}{10}
\providecommand{\url}[1]{#1}
\csname url@samestyle\endcsname
\providecommand{\newblock}{\relax}
\providecommand{\bibinfo}[2]{#2}
\providecommand{\BIBentrySTDinterwordspacing}{\spaceskip=0pt\relax}
\providecommand{\BIBentryALTinterwordstretchfactor}{4}
\providecommand{\BIBentryALTinterwordspacing}{\spaceskip=\fontdimen2\font plus
\BIBentryALTinterwordstretchfactor\fontdimen3\font minus
  \fontdimen4\font\relax}
\providecommand{\BIBforeignlanguage}[2]{{%
\expandafter\ifx\csname l@#1\endcsname\relax
\typeout{** WARNING: IEEEtran.bst: No hyphenation pattern has been}%
\typeout{** loaded for the language `#1'. Using the pattern for}%
\typeout{** the default language instead.}%
\else
\language=\csname l@#1\endcsname
\fi
#2}}
\providecommand{\BIBdecl}{\relax}
\BIBdecl

\bibitem{begoli2019need}
E.~Begoli, T.~Bhattacharya, and D.~Kusnezov, ``The need for uncertainty
  quantification in machine-assisted medical decision making,'' \emph{Nature
  Machine Intelligence}, vol.~1, no.~1, pp. 20--23, 2019.

\bibitem{wiens2019no}
J.~Wiens, S.~Saria, M.~Sendak, M.~Ghassemi, V.~X. Liu, F.~Doshi-Velez, K.~Jung,
  K.~Heller, D.~Kale, M.~Saeed \emph{et~al.}, ``Do no harm: a roadmap for
  responsible machine learning for health care,'' \emph{Nature medicine},
  vol.~25, no.~9, pp. 1337--1340, 2019.

\bibitem{qayyum2020secure}
A.~Qayyum, J.~Qadir, M.~Bilal, and A.~Al-Fuqaha, ``Secure and robust machine
  learning for healthcare: A survey,'' \emph{arXiv preprint arXiv:2001.08103},
  2020.

\bibitem{burton2020mind}
S.~Burton, I.~Habli, T.~Lawton, J.~McDermid, P.~Morgan, and Z.~Porter, ``Mind
  the gaps: Assuring the safety of autonomous systems from an engineering,
  ethical, and legal perspective,'' \emph{Artificial Intelligence}, vol. 279,
  p. 103201, 2020.

\bibitem{du2020ai}
X.~Du-Harpur, F.~Watt, N.~Luscombe, and M.~Lynch, ``What is ai? applications of
  artificial intelligence to dermatology,'' \emph{British Journal of
  Dermatology}, 2020.

\bibitem{sharkey2019autonomous}
A.~Sharkey, ``Autonomous weapons systems, killer robots and human dignity,''
  \emph{Ethics and Information Technology}, vol.~21, no.~2, pp. 75--87, 2019.

\bibitem{Davenport2019}
B.~E. M. A. J.~H. Davenport, T.~H. and R.~Wilson, \emph{Artificial
  Intelligence: The Insights You Need from Harvard Business Review}.\hskip 1em
  plus 0.5em minus 0.4em\relax Harvard Business Press, 2019.
  
  \bibitem{ISO_AI}
\BIBentryALTinterwordspacing
ISO, ``Iso/iec jtc 1/sc 42: Artificial intelligence,'' 2017, last accessed 10
  MAy 2020. [Online]. Available:
  \url{https://www.iso.org/committee/6794475.html}
\BIBentrySTDinterwordspacing

\bibitem{UK_GOV}
\BIBentryALTinterwordspacing
U.~C. on~Standards~in Public~Life, ``Artificial intelligence and public
  standards,'' 2020, last accessed 10 MAy 2020. [Online]. Available:
  \url{https://www.gov.uk/government/publications/artificial-intelligence-and-public-standards-report}
  
  \bibitem{alexander2020safety}
R.~Alexander, H.~Asgari, R.~Ashmore, A.~Banks, R.~Bongirwar, B.~Bradshaw,
  J.~Bragg, J.~Clegg, J.~Fenn, C.~Harper \emph{et~al.}, ``Safety assurance
  objectives for autonomous systems,'' 2020.



\bibitem{amodei2016concrete}
D.~Amodei, C.~Olah, J.~Steinhardt, P.~Christiano, J.~Schulman, and D.~Man{\'e},
  ``Concrete problems in ai safety,'' \emph{arXiv preprint arXiv:1606.06565},
  2016.
\bibitem{goodfellow2014generative}
I.~Goodfellow, J.~Pouget-Abadie, M.~Mirza, B.~Xu, D.~Warde-Farley, S.~Ozair,
  A.~Courville, and Y.~Bengio, ``Generative adversarial nets,'' in
  \emph{Advances in neural information processing systems}, 2014, pp.
  2672--2680.
  
  \bibitem{gulrajani2017improved}
I.~Gulrajani, F.~Ahmed, M.~Arjovsky, V.~Dumoulin, and A.~C. Courville,
  ``Improved training of wasserstein gans,'' in \emph{Advances in neural
  information processing systems}, 2017, pp. 5767--5777.



\BIBentrySTDinterwordspacing





\bibitem{bellemare2017cramer}
M.~G. Bellemare, I.~Danihelka, W.~Dabney, S.~Mohamed, B.~Lakshminarayanan,
  S.~Hoyer, and R.~Munos, ``The cramer distance as a solution to biased
  wasserstein gradients,'' \emph{arXiv preprint arXiv:1705.10743}, 2017.



\bibitem{van2014accelerating}
L.~Van Der~Maaten, ``Accelerating t-sne using tree-based algorithms,''
  \emph{The Journal of Machine Learning Research}, vol.~15, no.~1, pp.
  3221--3245, 2014.
  
  \bibitem{schulam2019can}
P.~Schulam and S.~Saria, ``Can you trust this prediction? auditing pointwise
  reliability after learning,'' \emph{arXiv preprint arXiv:1901.00403}, 2019.
  
  \bibitem{klas2019uncertainty}
M.~Kl{\"a}s and L.~Sembach, ``Uncertainty wrappers for data-driven models,'' in
  \emph{International Conference on Computer Safety, Reliability, and
  Security}.\hskip 1em plus 0.5em minus 0.4em\relax Springer, 2019, pp.
  358--364.





\bibitem{hobbhahn2020fast}
M.~Hobbhahn, A.~Kristiadi, and P.~Hennig, ``Fast predictive uncertainty for
  classification with bayesian deep networks,'' \emph{arXiv preprint
  arXiv:2003.01227}, 2020.
  
  \bibitem{quionero2009dataset}
J.~Quionero-Candela, M.~Sugiyama, A.~Schwaighofer, and N.~D. Lawrence,
  \emph{Dataset shift in machine learning}.\hskip 1em plus 0.5em minus
  0.4em\relax The MIT Press, 2009.

\bibitem{Theodoridis2009}
S.~Theodoridis and K.~Koutroumbas, \emph{Pattern Recognition}.\hskip 1em plus
  0.5em minus 0.4em\relax Elsevier Inc., 2009.

\bibitem{aslansefat2020performance}
K.~Aslansefat, M.~B. Gogani, S.~Kabir, M.~A. Shoorehdeli, and M.~Yari,
  ``Performance evaluation and design for variable threshold alarm systems
  through semi-markov process,'' \emph{ISA transactions}, vol.~97, pp.
  282--295, 2020.



\bibitem{fukunaga2013introduction}
K.~Fukunaga, \emph{Introduction to statistical pattern recognition}.\hskip 1em
  plus 0.5em minus 0.4em\relax Elsevier, 2013.

\bibitem{finlay2019empirical}
C.~Finlay and A.~M. Oberman, ``Empirical confidence estimates for
  classification by deep neural networks,'' \emph{arXiv preprint
  arXiv:1903.09215}, 2019.



\bibitem{cutillo2020machine}
C.~M. Cutillo, K.~R. Sharma, L.~Foschini, S.~Kundu, M.~Mackintosh, and K.~D.
  Mandl, ``Machine intelligence in healthcare—perspectives on
  trustworthiness, explainability, usability, and transparency,'' \emph{NPJ
  Digital Medicine}, vol.~3, no.~1, pp. 1--5, 2020.

\bibitem{nielsen2018chord}
F.~Nielsen, ``The chord gap divergence and a generalization of the
  bhattacharyya distance,'' in \emph{2018 IEEE International Conference on
  Acoustics, Speech and Signal Processing (ICASSP)}.\hskip 1em plus 0.5em minus
  0.4em\relax IEEE, 2018, pp. 2276--2280.

\bibitem{deza2014distances}
M.~M. Deza and E.~Deza, ``Distances in probability theory,'' in
  \emph{Encyclopedia of Distances}.\hskip 1em plus 0.5em minus 0.4em\relax
  Springer, 2014, pp. 257--272.

\bibitem{raschke2011empirical}
M.~Raschke, ``Empirical behaviour of tests for the beta distribution and their
  application in environmental research,'' \emph{Stochastic Environmental
  Research and Risk Assessment}, vol.~25, no.~1, pp. 79--89, 2011.

\bibitem{kabir2019runtime}
S.~Kabir, I.~Sorokos, K.~Aslansefat, Y.~Papadopoulos, Y.~Gheraibia, J.~Reich,
  M.~Saimler, and R.~Wei, ``A runtime safety analysis concept for open adaptive
  systems,'' in \emph{International Symposium on Model-Based Safety and
  Assessment}.\hskip 1em plus 0.5em minus 0.4em\relax Springer, 2019, pp.
  332--346.

\bibitem{papadopoulos2019model}
Y.~Papadopoulos, K.~Aslansefat, P.~Katsaros, and M.~Bozzano, \emph{Model-Based
  Safety and Assessment: 6th International Symposium, IMBSA 2019, Thessaloniki,
  Greece, October 16--18, 2019, Proceedings}.\hskip 1em plus 0.5em minus
  0.4em\relax Springer Nature, 2019, vol. 11842.

\bibitem{securitydataset}
I.~Sharafaldin, A.~H. Lashkari, and A.~A. Ghorbani, ``Toward generating a new
  intrusion detection dataset and intrusion traffic characterization.'' in
  \emph{International Conference on Information Systems Security and
  Privacy(ICISSP)}, 2018, pp. 108--116.

\bibitem{datasetanalysis}
R.~Panigrahi and S.~Borah, ``A detailed analysis of cicids2017 dataset for
  designing intrusion detection systems,'' \emph{International Journal of
  Engineering \& Technology}, vol.~7, no. 3.24, pp. 479--482, 2018.

\end{thebibliography}

\end{document}